\documentclass[prx,aps,twocolumn,amsmath,amssymb,longbibliography]{revtex4-1}

\pdfoutput=1
\usepackage{hyperref}
\usepackage{graphicx}
\usepackage{bm}
\usepackage[usenames]{color}
\usepackage{mathtools}

\DeclarePairedDelimiter{\floor}{\lfloor}{\floor}

\def\cal#1{\mathcal{#1}}
\def\eqq#1{Eq.~(\ref{#1})}

\def\eq#1{(\ref{#1})}
\def\f#1{Fig.~\ref{#1}}

\def\s#1{Section~\ref{#1}}
\def\c#1{~\cite{#1}}

\def\cc#1{Ref.~\cite{#1}}

\def\av#1{\langle #1 \rangle}

\def\x{\bm x}

\def\d{{\rm d}}
\def\u{U(\x)}

\def\beq{\begin{equation}}
\def\eeq{\end{equation}}
\def\bea{\begin{eqnarray}}
\def\eea{\end{eqnarray}}

\begin{document}

\title{Training neural networks using Metropolis Monte Carlo and an adaptive variant}

\author{Stephen Whitelam$^1$}
\email{swhitelam@lbl.gov}
\author{Viktor Selin$^2$}
\author{Ian Benlolo$^2$}
\author{Corneel Casert$^3$}
\author{Isaac Tamblyn$^{2,4}$}
\email{isaac.tamblyn@uottawa.ca}

\affiliation{$^1$Molecular Foundry, Lawrence Berkeley National Laboratory, 1 Cyclotron Road, Berkeley, CA 94720, USA\\
$^2$Department of Physics, University of Ottawa, ON, K1N 6N5, Canada \\   
$^3$Department of Physics and Astronomy, Ghent University, 9000 Ghent, Belgium \\   
$^4$Vector Institute for Artificial Intelligence, Toronto, ON M5G 1M1, Canada}

\begin{abstract}
We examine the zero-temperature Metropolis Monte Carlo algorithm as a tool for training a neural network by minimizing a loss function. We find that, as expected on theoretical grounds and shown empirically by other authors, Metropolis Monte Carlo can train a neural net with an accuracy comparable to that of gradient descent, if not necessarily as quickly. The Metropolis algorithm does not fail automatically when the number of parameters of a neural network is large. It can fail when a neural network's structure or neuron activations are strongly heterogenous, and we introduce an adaptive Monte Carlo algorithm, aMC, to overcome these limitations. The intrinsic stochasticity and numerical stability of the Monte Carlo method allow aMC to train deep neural networks and recurrent neural networks in which the gradient is too small or too large to allow training by gradient descent. Monte Carlo methods offer a complement to gradient-based methods for training neural networks, allowing access to a distinct set of network architectures and principles.
\end{abstract}
\maketitle

\section{Introduction}


The Metropolis Monte Carlo algorithm was developed in the 1950s in order to simulate molecular systems\c{metropolis1953equation,gubernatis2005marshall,rosenbluth2003genesis,whitacre2021arianna}. The Metropolis algorithm consists of small, random moves of particles, accepted probabilistically. It is, along with other Monte Carlo (MC) algorithms, widely used as a tool to equilibrate molecular systems\c{frenkel2001understanding}. Equilibrating a molecular system is similar in key respects to training a neural network: both involve optimizing quantities derived from many degrees of freedom that interact in a nonlinear way. Despite this similarity, the Metropolis algorithm and its variants are not widely used as a tool for training neural networks by minimizing a loss function (for exceptions, see e.g. Refs.\c{sexton1999beyond,rere2015simulated,tripathi2020rso}). Instead, this is usually done by gradient-based algorithms\c{schmidhuber2015deep,goodfellow2016deep}, and sometimes by population-based evolutionary or genetic algorithms\c{GA,GA2,montana1989training} to which Monte Carlo methods are conceptually related. 

In this paper we address the potential of the zero-temperature Metropolis Monte Carlo algorithm and an adaptive variant thereof as tools for neural-network training~\footnote{Zero temperature means that moves that increase the loss are not accepted. This choice is motivated by the empirical success in machine learning of gradient-descent methods, and by the intuition, derived from Gaussian random surfaces, that loss surfaces possess more downhill directions at large values of the loss\c{dauphin2014identifying,bahri2020statistical}.}. The class of algorithm we consider consists of taking a neural network of fixed structure, adding random numbers to all weights and biases simultaneously, and accepting this change if the loss function does not increase. For uncorrelated Gaussian random numbers this procedure is equivalent, for small updates, to normalized or clipped gradient descent in the presence of Gaussian white noise\c{kikuchi1991metropolis, kikuchi1992metropolis,whitelam2021correspondence}~\footnote{Note that algorithms of this nature do not constitute random search. The proposal step is random (related conceptually to the idea of weight guessing, a method used in the presence of vanishing gradients\c{hochreiter1997long}) but the acceptance criterion is a form of importance sampling, and leads to a dynamics equivalent to noisy gradient descent.}, and so its ability to train a neural network should be similar to that of simple gradient descent (GD). We show in \s{mnist} that, for a particular supervised-learning problem, this is the case, a finding consistent with results presented by other authors\c{sexton1999beyond,rere2015simulated,tripathi2020rso}.

It is sometimes stated that the ability of stochastic algorithms to train neural networks diminishes sharply as the number of network parameters increases (particularly if all network parameters are updated simultaneously). However, population-based evolutionary algorithms have been used to train many-parameter networks\c{salimans2017evolution}, and in \s{acc} we show that the ability of Metropolis Monte Carlo to train a fully-connected neural network is similar for networks with of order a hundred parameters or of order a million: there is not necessarily a sharp decline of acceptance rate with increasing network size. 

What {\em does} thwart the Metropolis Monte Carlo algorithm is network heterogeneity. For instance, if the number of connections entering neurons differs markedly throughout the network (as is the case for networks with convolutional- and fully-connected layers) or if the outputs of neurons in different parts of a network differ markedly (as is the case for very deep networks) then stochastic weight changes of a fixed scale will saturate neurons in some parts of the network and scarcely effect change in other parts. The result is an inability to train. To address this problem we introduce a set of simple adaptive modifications of the Metropolis algorithm -- a momentum-like term, an adaptive step-size scheduler, and a means of enacting heterogenous weight updates -- that are borrowed from ideas commonly used with gradient-based methods. The resulting algorithm, which we call adaptive Monte Carlo or aMC, is substantially more efficient than the non-adaptive Metropolis algorithm in a variety of settings. In \s{acc} we show, for one particular problem, that the acceptance rate of aMC remains much higher than that of the Metropolis algorithm at low values of loss, and can be made almost insensitive to network width, depth, and size. In \s{adaptive} we show that its momentum-like term speeds the rate at which aMC can learn the high-frequency features of an objective function, much as adaptive methods such as Adam\c{kingma2014adam} can learn high-frequency features faster than regular gradient descent. In \s{rnn} we show that the Monte Carlo method can train simple recurrent neural networks in the presence of small or large gradients, where gradient-based methods fail. In \s{deep} we show aMC can train deep neural networks in which gradients are too small for gradient-based methods to train. In \s{num} we comment on the fact that best practices for training nets using Monte Carlo methods await development. We introduce the elements of aMC throughout \s{results}, and summarize the algorithm in \s{amc}.

Our conclusion is that the Metropolis Monte Carlo algorithm and its adaptive variants such as aMC are viable ways of training neural networks, if less developed and optimized than modern gradient-based methods. In particular, Monte Carlo algorithms can, for small updates, effectively sense the gradient, and they do not fail simply because the number of parameters of a neural network becomes large. Monte Carlo algorithms should be considered a complement to gradient-based algorithms because they admit different design principles for neural networks. Given a network that permits gradient flow, modern gradient-based algorithms are fast and effective\c{schmidhuber2015deep, lecun2015deep,goodfellow2016deep}. For large neural nets with tens of millions of parameters we find gradient-based methods to be considerably faster than Monte Carlo (\s{num}). However, Monte Carlo algorithms free us from the requirement of ensuring reliable gradient flow (and gradients can be unreliable even in differentiable systems\c{metz2021gradients}). As a result, we find that Monte Carlo methods can train deep neural networks and simple recurrent neural networks in which gradients are too small (or too large) for gradient-based methods to work. There already exist solutions to these problems, namely the introduction of skip connections or more elaborate recurrent neural network architectures, but aMC requires neither of these things. One type of solution is architectural, the other algorithmic, and having both options offers more possibilities than having only one.

\section{Results}
\label{results}
\subsection{Metropolis Monte Carlo and its connection to gradient descent}
\label{mnist}

We start with the zero-temperature Metropolis Monte Carlo algorithm. The zero-temperature limit is not often used in molecular simulation, but it and its variants are widely used (and sometimes called random-mutation hill climbing) for optimizing non-differentiable systems such as cellular automata\c{mitchell1993will,mitchell1998introduction}. Consider a neural network with $N$ parameters (weights and biases) $\x=\{x_1,\dots,x_i,\dots,x_N\}$, and associated loss function $U(\x)$. If we propose the simultaneous change of each neural-network parameter by a Gaussian random number~\footnote{In Metropolis Monte Carlo simulations of molecular systems it is usual to propose moves of one particle at a time. If we consider neural-net parameters to be akin to particle coordinates then the analog would be to make changes to one neural-net parameter at a time; see e.g. Ref.\c{tripathi2020rso}. However, there is no formal mapping between particles and a neural network, and we could equally well consider the neural-net parameters to be akin to the coordinates of a single particle, in a high-dimensional space, in an external potential equal to the loss function. In the latter case the analog would be to propose a change of all neural-net parameters simultaneously, as we do here.},
\beq
\label{mutate}
x_i \to x_i + \epsilon_i \quad {\rm with}  \quad \epsilon_i \sim {\cal N}(0,\sigma^2),
\eeq
and accept the proposal if the loss does not increase, then, when the basic move scale $\sigma$ is small, the values $x_i$ of the neural-net parameters evolve according to the Langevin equation\c{whitelam2021correspondence}
\bea
\label{lang}
\frac{\d x_i}{\d n} = -\frac{\sigma}{\sqrt{2 \pi}}\frac{1}{|\nabla U(\x)|}\frac{\partial \u}{\partial x_i}+ \eta_i(n).
\eea
Here $n$ is training time (epoch), and $\eta$ is a Gaussian white noise with zero mean and variance $\av{\eta_i(n)\eta_j(n')}=(\sigma^2/2) \delta_{ij} \delta(n-n')$. That is, small stochastic perturbations of a network's weights and biases, accepted if the loss function does not increase, is equivalent to noisy clipped or normalized gradient descent on the loss function.

Given the success of gradient-based training methods, this correspondence shows the potential of the Metropolis algorithm to train neural networks. Consistent with this expectation, we show in \f{fig_zero} that the zero-temperature Metropolis algorithm can train a neural network. For comparison, we also train the network using simple gradient descent, 
\beq
\label{gd}
\x \to \x  -\alpha \nabla U(\x),
\eeq
where $\alpha$ is the learning rate, $U(\x)$ the loss function, and $\nabla \equiv (\partial/\partial x_1,\dots,\partial/\partial x_N)$ the gradient operator with respect to the neural-network parameters $\x$.
\begin{figure}[] 
   \centering
   \includegraphics[width=\linewidth]{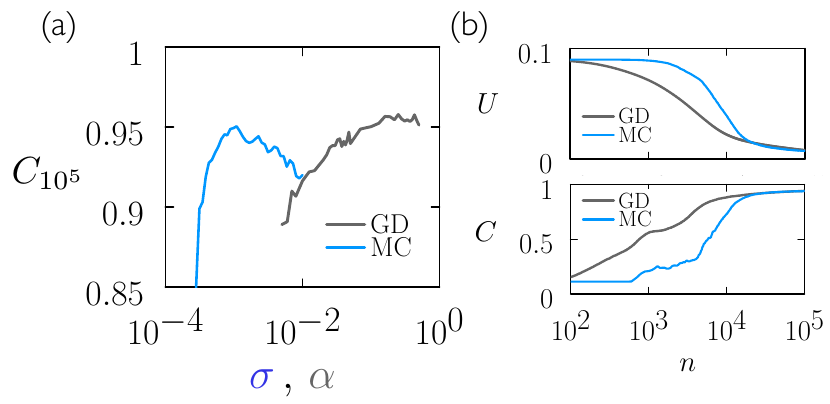} 
   \caption{Comparison of zero-temperature Metropolis Monte Carlo (MC) and gradient descent (GD) used to train a neural network to minimize the mean-squared error $U$ on the MNIST training set. (a) MNIST test-set accuracy $C_{10^5}$ after $10^5$ epochs of batch learning as a function of MC step size $\sigma$ (blue) or GD learning rate $\alpha$ (gray). (b) Training loss $U$ and test-set accuracy $C$ versus epoch $n$ for MC step size $\sigma=2 \times 10^{-3}$ (blue) and GD learning rate $\alpha = 4.5 \times 10^{-2}$ (gray). }
   \label{fig_zero}
\end{figure}

We consider a standard supervised-learning problem, recognizing MNIST images\c{mnist,lecun1998gradient} using a fully-connected, two-layer neural net. The net has 784 input neurons, 16 neurons in each hidden layer, and one output layer of 10 neurons. The hidden neurons have hyperbolic tangent activation functions, and the output neurons comprise a softmax function. The net has in total 13,002 parameters\c{nn}. We did batch learning, with the loss function $U$ being the mean-squared error on the MNIST training set of size $6\times 10^4$ (in the standard way we consider the ground truth for each training example to be a 1-hot encoding of the class label, and take the 10 outputs of the neural network as its prediction vector).

\f{fig_zero}(a) shows the classification accuracy $C_{10^5}$ on the MNIST test set of size $10^4$ after $10^5$ epochs of training. We show results for MC (blue) and GD (gray), for a range of values of step size $\sigma$ and learning rate $\alpha$, respectively. The initial neural-net parameters for MC simulations were $x_i\sim {\cal N}(0,\sigma^2)$. The two algorithms behave in a similar manner: each has a range of its single parameter over which it is effective, and displays a maximum at a particular value of that parameter. The value of the maximum for GD is slightly higher than that for MC (about 96\% compared to about 95\%), and GD achieves near-maximal results over a broader range of its single parameter than does MC.
\begin{figure*}[] 
   \centering
   \includegraphics[width=\linewidth]{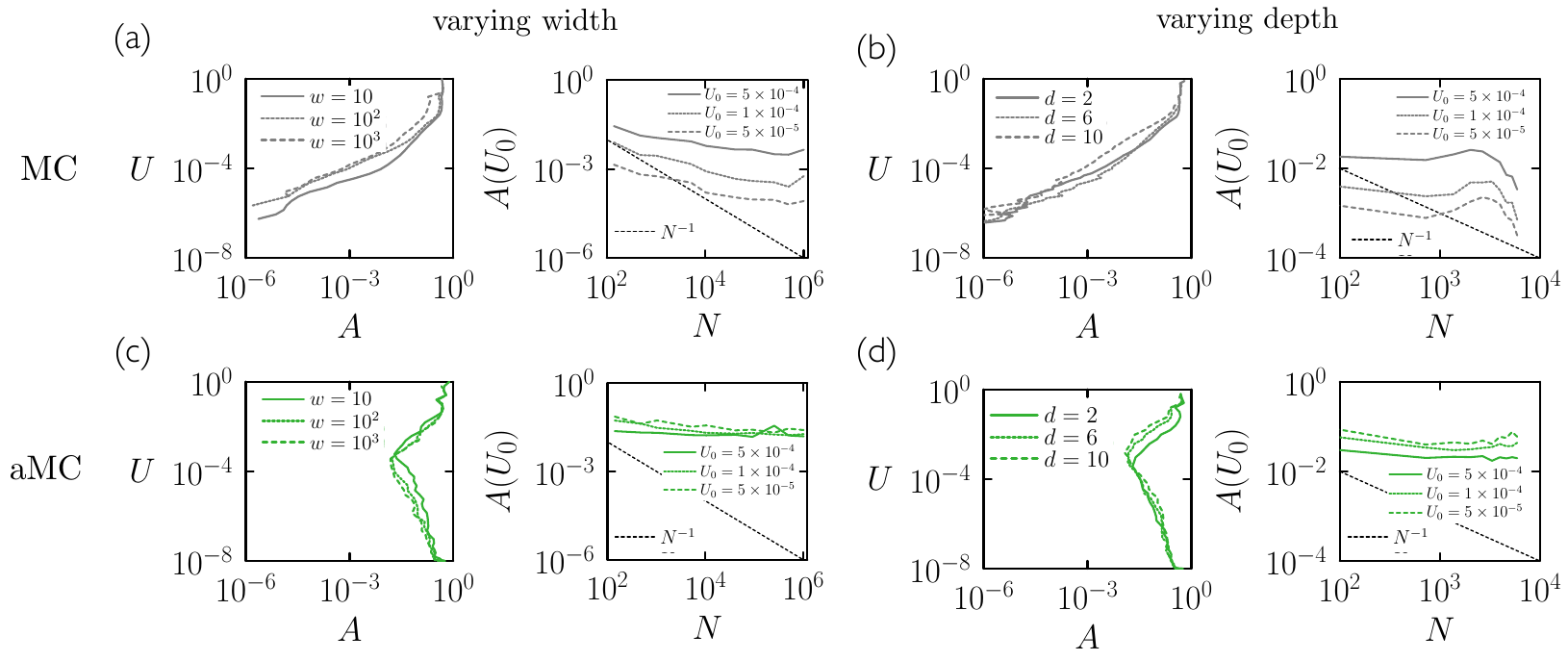} 
   \caption{For a particular neural-network supervised-learning task, the acceptance rate of the Metropolis MC method decreases sharply with loss but not model size. (a) Left: Loss $U$ versus acceptance rate $A$ for Metropolis MC for a deep neural net of three different widths. Right: Acceptance rate $A(U_0)$ at fixed loss $U_0$ for models of different widths, as a function of the number of model parameters $N$. (b) Analog of (a) for models of increasing depth. (c) Analog of (a) for adaptive Monte Carlo, aMC. (d) Analog of (b) for aMC. MC step size: $\sigma=10^{-3}$. aMC hyperparameters: $(\sigma_0,\epsilon,n_{\rm s}, {\rm signal}\, {\rm norm})=(10^{-2},10^{-2},10^2, {\rm on})$. Each data point is the average of 5 independent simulations.}
   \label{fig_acc}
   \end{figure*}
   
\f{fig_zero}(b) shows loss $U$ and classification accuracy $C$ as a function of epoch for two examples from panel (a). GD trains faster with $n$ initially, but results are comparable near the end of the learning process. The learning dynamics of these algorithms is not the same: in the limit of small steps, the zero-temperature Metropolis algorithm approaches normalized or clipped gradient descent, not standard gradient descent (and its equivalence to the former would only be seen with an appropriately rescaled horizontal axis)~\footnote{We have also found the GD-MC equivalence to break down in other circumstances: for certain learning rates $\alpha$, the discrete-update equation \eq{gd} sometimes results in moves uphill in loss, in which case the discrete update is not equivalent to the equation $\dot{\x} = -(\alpha/\Delta t) \nabla U(\x)$, while the latter {\em is} equivalent to the small-step-size limit of the finite-temperature Metropolis algorithm\c{kikuchi1991metropolis, kikuchi1992metropolis,whitelam2021correspondence}.}. Nonetheless, MC can in effect sense the gradient, as long as the step size is relatively small, and for this problem the range of appropriate step sizes is small compared to the effective GD step size. GD therefore trains faster, but MC has similar capacity for learning. The computational cost per epoch of the two algorithms is of similar order, with MC being cheaper per epoch for batch learning: each MC step requires a forward pass through the data, and each GD step a forward and a backward pass.

There are many things that could be done to improve the learning precision of these algorithms (no pre-processing of data was done, and a basic neural net was used~\footnote{In \f{fig_resnet} we show that GD and MC can both train a large modern neural network to a classification accuracy in excess of 99\% on the same problem.}), but this comparison, given a neural net and a data set, confirms that Metropolis Monte Carlo can achieve results roughly comparable to gradient descent, even on a problem for which gradients are available. For this problem GD trains faster, but MC works. It is worth noting that this conclusion follows from considering a range of step sizes $\sigma$: for a single choice of step size it would be possible to conclude that MC doesn't work at all.
 
Similar findings have been noted previously: simulated annealing on the Metropolis algorithm\c{sexton1999beyond,rere2015simulated} and a variant of zero-temperature Metropolis MC (applied weight-by-weight)\c{tripathi2020rso} were used to train neural nets with an accuracy comparable to that of gradient-based algorithms. These results, and the correspondence described in~\cc{whitelam2021correspondence} establish both theoretically and empirically the ability of Metropolis MC to train neural nets. 

We next turn to the question of how the efficiency of Monte Carlo training scales with net parameters, and how to improve this efficiency by introducing adaptivity to the algorithm.

\subsection{Metropolis acceptance rate as a function of net size}
\label{acc}

To examine how the efficiency of the Metropolis algorithm changes with neural-net size and architecture, we consider in this section a supervised-learning problem in which a neural net is trained by zero-temperature Metropolis MC to express the sine function $f_0(\theta) = \sin (2\pi \theta)$ on the interval $\theta \in[0,1]$. The loss $U$ is the mean-squared error between $f_0(\theta)$ and the neural-net output $f_{\bm x}(\theta)$, evaluated at $10^3$ evenly-spaced points on the interval. The neural net has one input neuron, which is fed the value $\theta $, and one output, which returns $f_{\bm x}(\theta)$. Its internal structure is fully connected, with hyperbolic tangent nonlinearities. To explore the effect of varying width (panels (a) and (c) of \f{fig_acc}) we set the depth to 2 and varied the width from $10$ to $10^3$, these choices corresponding to about $10^2$ to about $10^6$ parameters. To explore the effect of varying depth (panels (b) and (d) of \f{fig_acc}) we set the width to 25 and varied the depth from 2 to 10, these choices corresponding to about $10^2$ to about $10^4$ parameters. 

In \f{fig_acc}(a,left) we show loss $U$ as a function of Metropolis acceptance rate $A$ for three different neural-net widths. The acceptance rate tells us, for fixed step size, the fraction of directions that point downhill in loss. It provides information similar to that shown in plots of the index of the critical points of a loss surface\c{dauphin2014identifying,bahri2020statistical}, confirming that at large loss there are more downhill directions than at small loss. In \f{fig_acc}(a,right) we plot the acceptance rate $A(U_0)$ at fixed loss $U_0$ as a function of the number of net parameters $N$ (obtained by taking horizontal cuts across panel (a); note that more net sizes are shown in panel (b) than panel (a)). 

The acceptance rate decreases with increasing net size, but relatively slowly. Upon increasing the size of the net by 4 orders of magnitude, the acceptance rate decreases by about 1 order of magnitude. We have indicated an $N^{-1}$ scaling as a guide to the eye. In the extreme limit, if $N$ simultaneous parameter updates each had to be individually productive, the acceptance rate would decrease exponentially with $N$, which is clearly not the case. The more dramatic decrease in acceptance rate is with loss: at small loss the acceptance rate becomes very small. 

Similar trends are seen with depth in \f{fig_acc}(b). The acceptance rate declines sharply with loss. It also declines with the number of parameters, slightly more rapidly than in panel (a) but not as rapidly as $N^{-1}$.

Empirically, therefore, we do not see evidence of a fundamental inability of MC to cope with large numbers of parameters. In \s{deep} we discuss how network heterogeneity can impair the Metropolis algorithm's ability to train a network. The solution, as we discuss there, is to introduce an adaptive Monte Carlo (aMC) variant of the Metropolis algorithm. To motivate the introduction of this algorithm we show in panels (c) and (d) the aMC analog of panels (a) and (b), respectively. The trends experienced by Metropolis have been annulled, the acceptance rates of aMC remaining large and essentially constant with loss or model size over the range of parameters considered. 

We now turn to a step-by-step introduction of the elements of aMC.

\subsection{Adaptivity speeds learning, particularly of high-frequency features}
\label{adaptive}

Modern gradient-based methods are adaptive, allowing the learning rate for each neural-net parameter to differ and to change as a function of the gradients encountered during training\c{goodfellow2016deep} (adaptive learning is also used in evolutionary algorithms\c{hansen2001completely,hansen2003reducing,hansen2006cma}). We can copy this general idea in a simple way within a zero-temperature Metropolis Monte Carlo scheme by changing the proposed move of \eqq{mutate} to 
\beq
\label{step2}
\epsilon_i \sim {\cal N}(\mu_i,\sigma^2).
\eeq 
The parameters $\mu_i$, set initially to zero, are updated after every accepted move according to
\beq
\mu_i \to \mu_i + \epsilon(\epsilon_i-\mu_i),
\eeq
where $\epsilon$ is a hyperparameter of the method. This form of adaptation is similar to the inclusion of momentum in a gradient-based scheme: the center $\mu_i$ of each parameter's move-proposal distribution shifts in the direction of the last accepted move $\epsilon_i$, with the aim of increasing the probability of generating moves that will be accepted. In addition, to remove the need for a search over the step-size parameter $\sigma$, we introduce a simple adaptive learning-rate schedule by setting $\sigma \to 0.95 \sigma$ (and $\mu_i=0$) after $n_{\rm s}$ consecutive rejected moves. We take the initial step size to be $\sigma=\sigma_0$. We refer to this adaptive Monte Carlo algorithm as aMC, specified in \s{amc}.

\begin{figure}[] 
   \centering
   \includegraphics[width=0.7\linewidth]{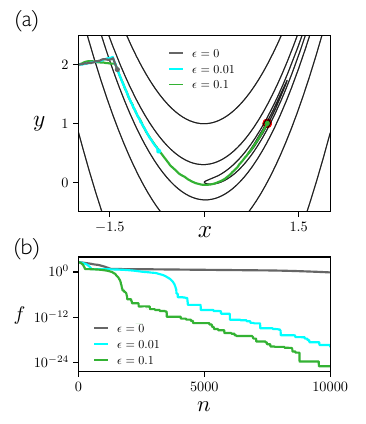} 
   \caption{Adaptivity speeds convergence on a test function. We use aMC with three values of $\epsilon$ to simulate a particle moving on the Rosenbrock function $f(x,y)$, \eqq{banana}. Panel (a) shows a 2000-step simulation for each of three values of $\epsilon$; the global minimum is indicated by the red dot at (1,1). Panel (b) shows the value of the function along each trajectory when continued to $n=10^4$ steps. aMC hyperparameters: $\sigma_0=10^{-3}, n_{\rm s}=20$.}
   \label{fig_banana}
\end{figure}

The aMC parameter $\epsilon$ can be used to influence the rate of learning. In \f{fig_banana} we provide a simple illustration of this fact using the two-dimensional Rosenbrock function
\beq
\label{banana}
f(x,y) = (1-x)^2 + 100(y-x^2)^2,
\eeq
often used as a test function for optimization methods\c{rosenbrock1960automatic,shang2006note,emiola2021comparison}. The Rosenbrock function has a global minimum at $(x,y) = (1,1)$ that is set within a long valley surrounded by steep slopes on either side. A particle on this function moving under pure gradient descent takes a long time to reach the global minimum because gradients within the valley are small: the particle will quickly reach the valley and then move slowly along the valley floor\c{rosenbrock1960automatic,shang2006note}. Placing a particle at the point $(-2,2)$, outside the valley, we evolve the position $\x=(x,y)$ of the particle using aMC. We used the initial step size $\sigma_0=10^{-3}$ and rescaling parameter $n_{\rm s}=100$, and carried out three simulations for three values of $\epsilon$. As shown in the figure, the larger is $\epsilon$ the more rapidly is the global minimum attained, with the difference between zero and nonzero epsilon being considerable. This form of adaptivity has an effect similar to that of momentum with gradient descent\c{goh2017momentum}.

Returning to neural-network simulation, the aMC parameter $\epsilon$ can speed the rate at which a neural network can learn high-frequency features, much as adaptive methods do with gradient-based algorithms. The extent to which high-frequency features {\em should} be learned varies by application. For instance, if a training set contains high-frequency noise then we may wish to attenuate an algorithm's ability to learn this noise in order to enhance its ability to generalize. This is the idea expressed in Fig. 2 of~\cc{rumelhart1995backpropagation}. Empirical studies show that non-adaptive versions of gradient descent sometimes generalize better than their adaptive counterparts\c{chen2018closing}, in some cases because of the different abilities of these things to learn high-frequency features. 

In \f{fig_gen} we consider a supervised-learning problem inspired by Fig. 2 of~\cc{rumelhart1995backpropagation}. We use gradient-based methods (GD and Adam) and aMC to train a neural network to learn the function 
\beq
\label{one}
f_0(\theta) = \ln(1+5 \theta) + \frac{1}{10}\sin(20 \pi \theta)
\eeq
on $\theta \in [0,1]$. This function contains a low-frequency term, the logarithm, and a high-frequency term, the sine. The neural network has one input neuron, which is fed the value $\theta$, one output neuron, which returns $f_{\bm x}(\theta)$, and a single hidden layer of 100 neurons with tanh activations. The parameters $\x$ of the network were set initially to random values $x_i \sim {\cal N}(0, \sigma_0^2)$. 
\begin{figure}[] 
   \centering
   \includegraphics[width=\linewidth]{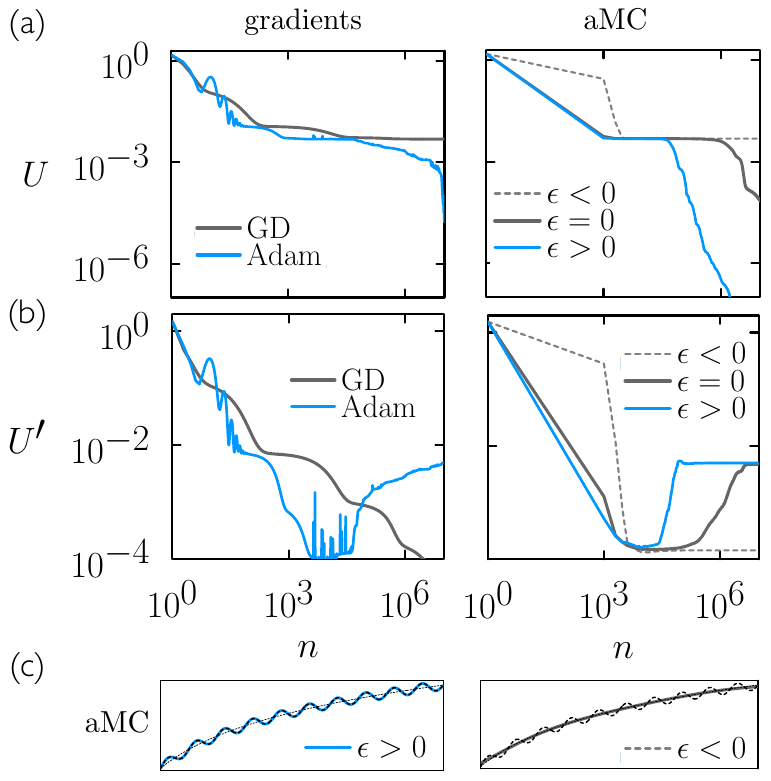} 
   \caption{Adaptivity speeds learning of high-frequency features. We train a neural network to learn the function \eq{one} using two gradient-based methods (GD and Adam) and aMC with three values of the parameter $\epsilon$. (a) Loss $U$ as a function of training time $n$. (b) Pseudo-loss $U'$, which expresses the difference between the net function and the low-frequency term (the logarithm) of \eq{one}. (c) Output of neural nets trained by aMC (colored lines) at fixed training time. The thicker and thinner black lines are the function \eq{one} and its low-frequency term, respectively. GD and Adam learning rate: $\alpha = 5 \times 10^{-3}$; aMC hyperparameters: $(\sigma_0,n_{\rm s}) = (0.1,50)$.}
   \label{fig_gen}
\end{figure}

In \f{fig_gen}(a) we show the training loss, the mean-squared difference $U$ between $f_0(\theta)$ and the neural-net output $f_{\bm x}(\theta)$, evaluated at 1000 evenly-spaced points over the interval. At left we show results produced using GD and Adam, and at right we show results produced using aMC for three values of $\epsilon$, one positive ($\epsilon=5 \times 10^{-3}$), one negative ($\epsilon=-5 \times 10^{-3}$), and zero. Of the gradient-based methods Adam trains faster than GD, while for aMC the training loss $U$ decreases fastest for positive $\epsilon$ and slowest for negative $\epsilon$. (We did not carry out a systematic search of learning rates in order to compare directly the gradient-based and Monte Carlo methods; our intent here is to illustrate how adaptivity matters within the two classes of algorithm.) 

In \f{fig_gen}(b) we show the pseudo-loss $U'$ that expresses the mean-squared difference between the net function $f_{\bm x}(\theta)$ and the low-frequency logarithmic term of $f_0$. The net is not trained to minimize $U'$, but it so happens during training that $U'$ becomes small as the net first learns the low-frequency component of $f_0$. Subsequently, $U'$ increases as the net also learns the high-frequency component of $f_0$. 
\begin{figure*}[] 
   \centering
   \includegraphics[width=\linewidth]{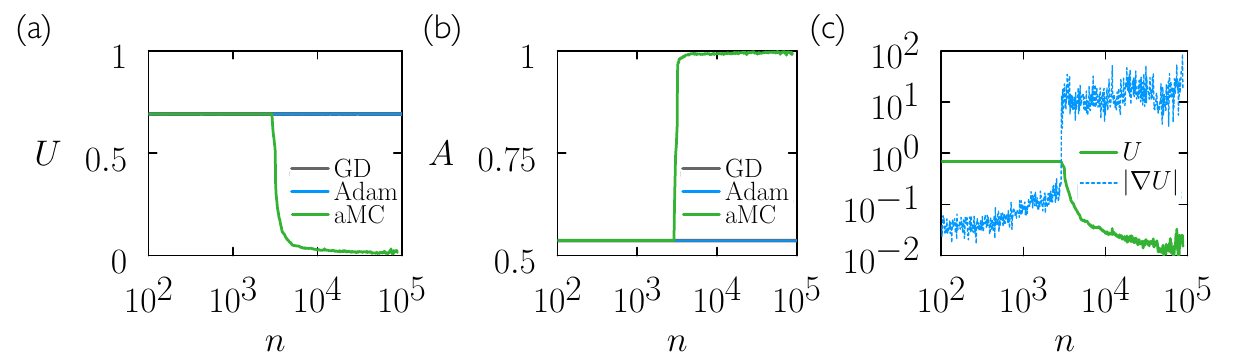} 
   \caption{The intrinsic stochasticity and numerical stability of the Monte Carlo method allows it to learn in the presence of numerically small and large gradients. (a) Training-set loss $U$ and (b) test-set classification accuracy $C$ as a function of training epoch $n$ for a simple RNN instructed to distinguish MNIST 0s from 1s. The RNN is presented with each image as an unrolled sequence of length 784. The gradient-based methods GD (gray) and Adam (blue) cannot train; aMC (green) trains to an accuracy of 99.8\%. (c) Size of the gradient (blue) associated with the models produced by aMC. The algorithm trains productively in regimes in which $|\nabla U| \ll U$ and $|\nabla U| \gg U$. The loss (green) is reproduced from panel (a). Learning rate for GD and Adam: $\alpha =10^{-3}$; aMC hyperparameters: $\sigma_0=10^{-3},n_{\rm s}=100, \epsilon=0$.}
   \label{fig_rnn}
\end{figure*}

Of the gradient-based methods Adam learns high-frequency features more rapidly than GD. As it does so, the value of the pseudo-loss $U'$ increases. For aMC, the parameter $\epsilon$ controls the separation of timescales between the learning of the low- and high-frequency components of $f_0$. If we want to learn $f_0$ as quickly as possible then positive $\epsilon$ is the best choice. But if we consider the high-frequency component of $f_0$ to be noise, and regard $U'$ as a measure of the network's generalization error, then negative $\epsilon$ is the best choice. 

Panel (c) shows the aMC net functions at a time $n$ such that the net trained using $\epsilon=0$ has begun to learn the high-frequency features of $f_0$. At the same time the nets trained using positive and negative $\epsilon$ have learned these features completely or not at all, respectively.

\subsection{Monte Carlo algorithms can train a neural network even when gradients are unreliable}
\label{rnn}

Metropolis Monte Carlo and aMC can effectively sense the gradient\c{whitelam2021correspondence}, and so can train a neural network to similar levels of accuracy as gradient descent. However, Monte Carlo algorithms can also train a neural network when gradients become unreliable, such as when they vanish or explode. 

Vanishing gradients can be overcome by the intrinsic stochasticity of the Monte Carlo method. In the absence of gradients, pure gradient-based methods receive no signal\c{hochreiter1997long,hochreiter1998vanishing}. However, the Metropolis Monte Carlo procedure \eq{mutate} is equivalent, for vanishing gradients, to the diffusive dynamics $\dot{x}_i= \xi_i(n)$, where $\xi$ is a Gaussian white noise with zero mean and variance $\av{\xi_i(n)\xi_j(n')}=\sigma^2 \delta_{ij} \delta(n-n')$. Thus Metropolis MC will, in the absence of gradients, enact diffusion in parameter space until nonvanishing gradients are encountered, at which point learning can resume. 

Monte Carlo algorithms can also cope with exploding gradients. Moves are proposed without reference to the gradient, and so can be made on a landscape for which the gradient varies rapidly or is numerically large. Monte Carlo algorithms are also numerically stable, with moves that would induce large increases in loss being rejected but otherwise not harming the training process.
\begin{figure*}[] 
   \centering
   \includegraphics[width=\linewidth]{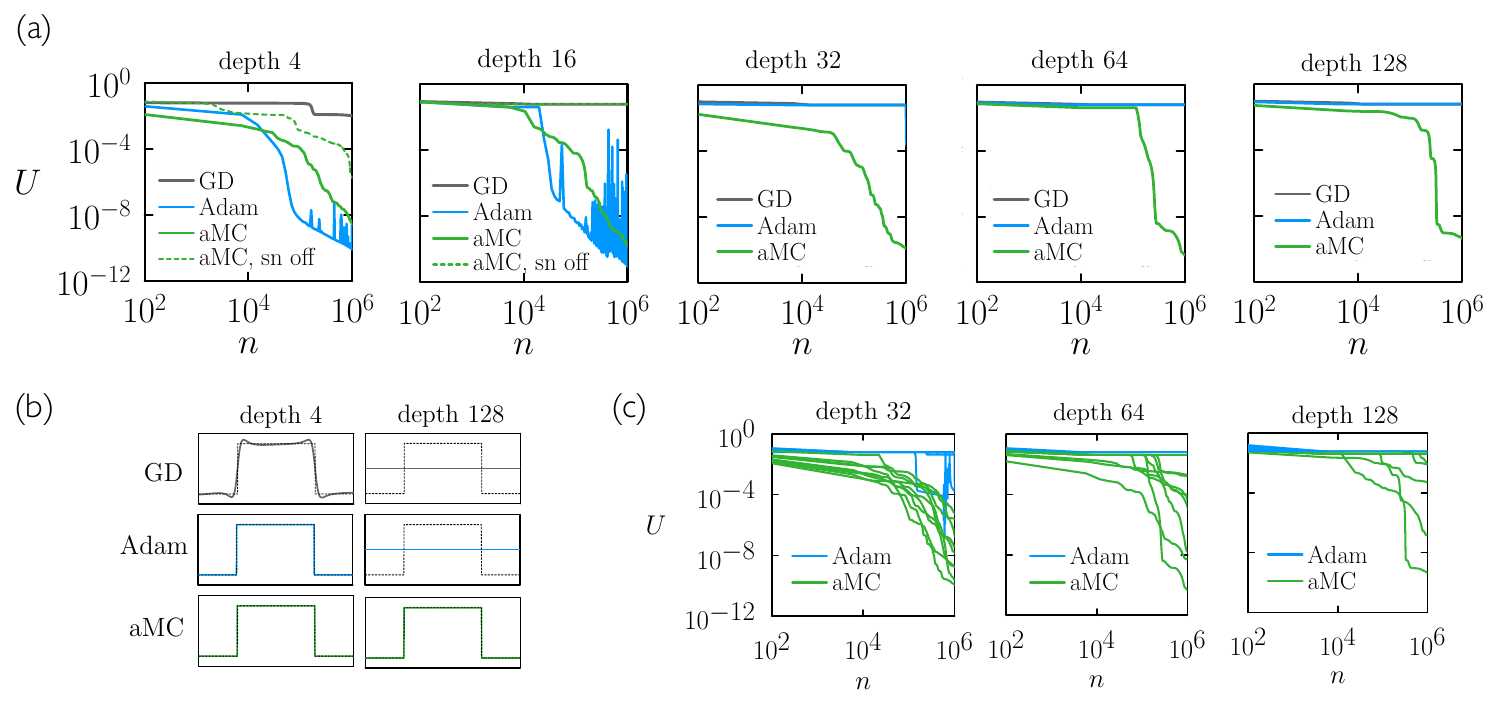} 
   \caption{Intrinsic stochasticity and heterogenous weight-scale updates allow aMC to train deep nets in which the gradient is too small for gradient-based methods to work. (a) Loss function $U$ versus epoch $n$ for deep nets trained to express a step function by GD (gray), Adam (blue), and aMC with signal norm off (green dashed) and on (green). The latter is able to train for all depths shown. (b) Net outputs after $10^6$ epochs (colored) against the target function (black) for two depths. (c) Loss $U$ of 10 independent simulations for each of Adam and aMC for the deeper nets. GD learning rate: $\alpha=10^{-3}$; Adam hyperparameters: $(\alpha, \beta_1, \beta_2) = (10^{-4},0.9,0.999)$; aMC hyperparameters: ($\sigma_0$,$\epsilon$,$n_{\rm s}$) = $(10^{-2},10^{-2},10^2)$.}
   \label{fig2}
\end{figure*}

We show in this section that the ability to cope with vanishing and exploding gradients allows aMC to train recurrent neural networks that gradient-based methods cannot. Recurrent neural networks (RNNs) are designed to act on sequences, and have been applied to problems involving text, images, speech, and music\c{medsker2001recurrent,graves2013speech,sutskever2014sequence,graves2008offline}.  An RNN possesses a vector known as its hidden state, which acts as a form of memory. This vector is updated each time the RNN views a position in a sequence. The size of the gradient scales in general exponentially with sequence length, and so when the sequence is long, and long-term dependencies exist, the gradient tends to vanish or explode. As a result, it can be difficult to train simple RNNs using first-order gradient-based methods in order to learn long-term dependencies in sequence data\c{schmidhuber2015deep,wierstra2010recurrent,bengio1994learning,martens2011learning,bengio2013advances}. One solution to this problem is the introduction of more elaborate and computationally expensive RNN architectures such as long short-term memory\c{hochreiter1997long} and gated recurrent units\c{cho2014learning,kanai2017preventing}. These architectures can be more reliably trained by gradient-based methods than can simple RNNs.

Here we demonstrate an algorithmic solution to the problem rather than an architectural one, by showing that Monte Carlo methods can train an RNN in circumstances in which gradient-based methods cannot. We train a simple RNN with tanh activation functions to distinguish the digits of class 0 and 1 in the MNIST data set. We binarized the data, and used a vector of length 2 and a one-hot encoding to represent black or white pixels. The RNN was shown an unrolled version of each image, a sequence of 784 pixels. The RNN architecture is a two-layer stack in which the hidden state at each site in the first layer is used as input for the second layer. The dimension of the hidden state is set to 64, and the final hidden state is sent into a linear classifier. The loss function is the cross entropy between the neural network's prediction and the ground truth. 

In \f{fig_rnn}(a,b) we show the results of training the RNN using aMC and gradient-based optimizers. We stochastically subsample the training data set for each update (``mini-batching"), using 1500 samples at each step. We used gradient descent and the Adam optimizer, both with learning rate $10^{-3}$, with gradient clipping turned on (this has been shown to solve the exploding-gradient problem for some data sets\c{pascanu2013difficulty}), and aMC with hyperparameters $\sigma_0=10^{-3},n_{\rm s}=100, \epsilon=0$ (which is the Metropolis algorithm with an adaptive step-size scheduler). Gradient descent and Adam fail to train (a search over learning rates between the values $10^{-5}$ and $1$ did not result in lower loss values), while aMC trains to small values of loss and a test-set accuracy of $99.8\%$.

In \f{fig_rnn}(c) we show the size of the gradient associated with the models produced by aMC. aMC does not calculate or make use of the gradient, but we can nonetheless evaluate it for the models produced by the training process. Two distinct regimes can be seen, with the size of the gradient changing by about three orders of magnitude between them. The gradient-based algorithms cannot escape from the small-gradient regime, and when initiated from the large-gradient regime that is encountered by aMC, in which $|\nabla U| \gg U$, the gradient-based integrators explode. Thus Monte Carlo can train productively in the face of two of the classic obstacles to training by gradient-based methods, small and large gradients. 

Simple RNNs are known be harder to train by gradient descent than more complex RNNs, but simple RNNs possess similar or greater {\em capacity} per parameter than more complex architectures\c{collins2016capacity}. Methods that can train simple RNNs may therefore allow more widespread use of those architectures.

\subsection{For nets with heterogeneous structures or neuron activations, the weight-update scale must be made heterogenous}
\label{deep}

For some architectures, particularly those with structural heterogeneity or heterogenous neuron activations, it is necessary to scale the Monte Carlo step-size parameter $\sigma$ for each neural-net parameter individually. In this section we address this problem using deep neural networks for which, as in \s{rnn}, gradients are too small for gradient-based methods to train.

Choosing a set of heterogeneous Monte Carlo step-size parameters can be done by adapting ideas used in the development of gradient-based methods\c{lecun1996effiicient}. Guided by that work we modify the proposal distribution of \eq{step2} to read
\beq
\label{step3}
\epsilon_i \sim {\cal N}(\mu_i,\sigma_i^2),
\eeq 
where $\sigma_i = \lambda_i \sigma$. The $\lambda_i$ are parameters that are either set to unity (a condition we call ``signal norm off'') or according to \eqq{fan_in} in \s{sn} (``signal norm on''). The parameters $\mu_i$ and $\sigma$ are adjusted as in \s{adaptive}. The parameters $\lambda_i$, which are straightforwardly calculated during a forward pass through the net, ensure that the scale of signal change to each neuron is roughly constant. The intent of signal norm is similar to that of layer norm\c{ba2016layer}, except that the latter is an architectural solution -- it entails a modification of the net, and is present at test time -- while the former is an algorithmic solution and plays no role once the neural net has been trained.

In \f{fig2} we show the results of neural networks of depth $d$ trained by aMC and by gradient-based methods to express a step function $f_0(\theta)$ that is equal to 1/2 if $1/2< \theta < 3/4$ and is zero otherwise. The neural nets have one input neuron, which is fed the value $\theta$, and one output neuron, which returns $f_{\bm x}(\theta)$. They have 10 neurons in the penultimate hidden layer, and 4 neurons in each of the other $d-1$ hidden layers, the intent being to allow very deep nets with relatively few neurons. All neurons have tanh activation functions. 

In \f{fig2}(a) we show loss $U$ as a function of epoch $n$ for four algorithms: GD (gray); Adam (blue); and aMC with signal norm off (green dotted) and on (green). For each algorithm we ran 20 independent simulations, 10 using Kaiming initialization\c{paszke2019pytorch} and 10 initialized with Gaussian random numbers $x_i \sim {\cal N}(0 ,\sigma_0^2)$, where $\sigma_0 = 10^{-2}$. We plot the simulation having the smallest $U$ after $10^6$ epochs. As the depth of the network increases beyond 4 layers, GD and aMC with signal norm off stop learning on the timescales shown. Above 32 layers, Adam also stops learning on the timescale shown. (For depth 64 we tried a broad range of learning rates for GD and Adam, from $10$ to $10^{-6}$, none of which was successful. We also varied the Adam hyperparameters $\beta_1, \beta_2$ over a small range of values, without success. It may be that hyperparameters that enable training do exist, but we were not able to find them.) aMC with signal norm on continues to learn up to a depth of 128 (we also verified that aMC trains nets of depth 256), and so can successfully train deep nets in which gradient-based algorithms receive too little signal to train. 

In \f{fig2}(b) we show net outputs at $n=10^6$ epochs for three algorithms and two depths. As discussed in \s{adaptive}, the adaptive algorithms Adam and aMC learn the sharp features of the target function more quickly than does GD. For the deeper net, the gradient-based algorithms GD and Adam do not receive sufficient signal to train.

In \f{fig2}(c) we show 10 simulations for each of Adam and aMC for the deeper nets. The outcome of training is stochastic, resembling a nucleation dynamics with an induction time that increases with net depth. For some initial conditions both algorithms fail to train on the allotted timescale. In general, the rate of nucleation is higher for aMC than Adam, and remains measurable for all depths shown

We verified that introducing skip connections\c{he2016deep} or layer norm\c{ba2016layer} to the neural nets enabled Adam to train at depth 64 (and enables aMC without signal norm to train at all depths). These architectural modifications allow gradient-based algorithms to train, but they are not required for Monte Carlo. This comparison illustrates the fact that design principles for nets trained by Monte Carlo methods differ from those trained by gradient-based methods.

\subsection{Best practices for training neural nets using MC await development}
\label{num}
\begin{figure}[] 
   \centering
   \includegraphics[width=\linewidth]{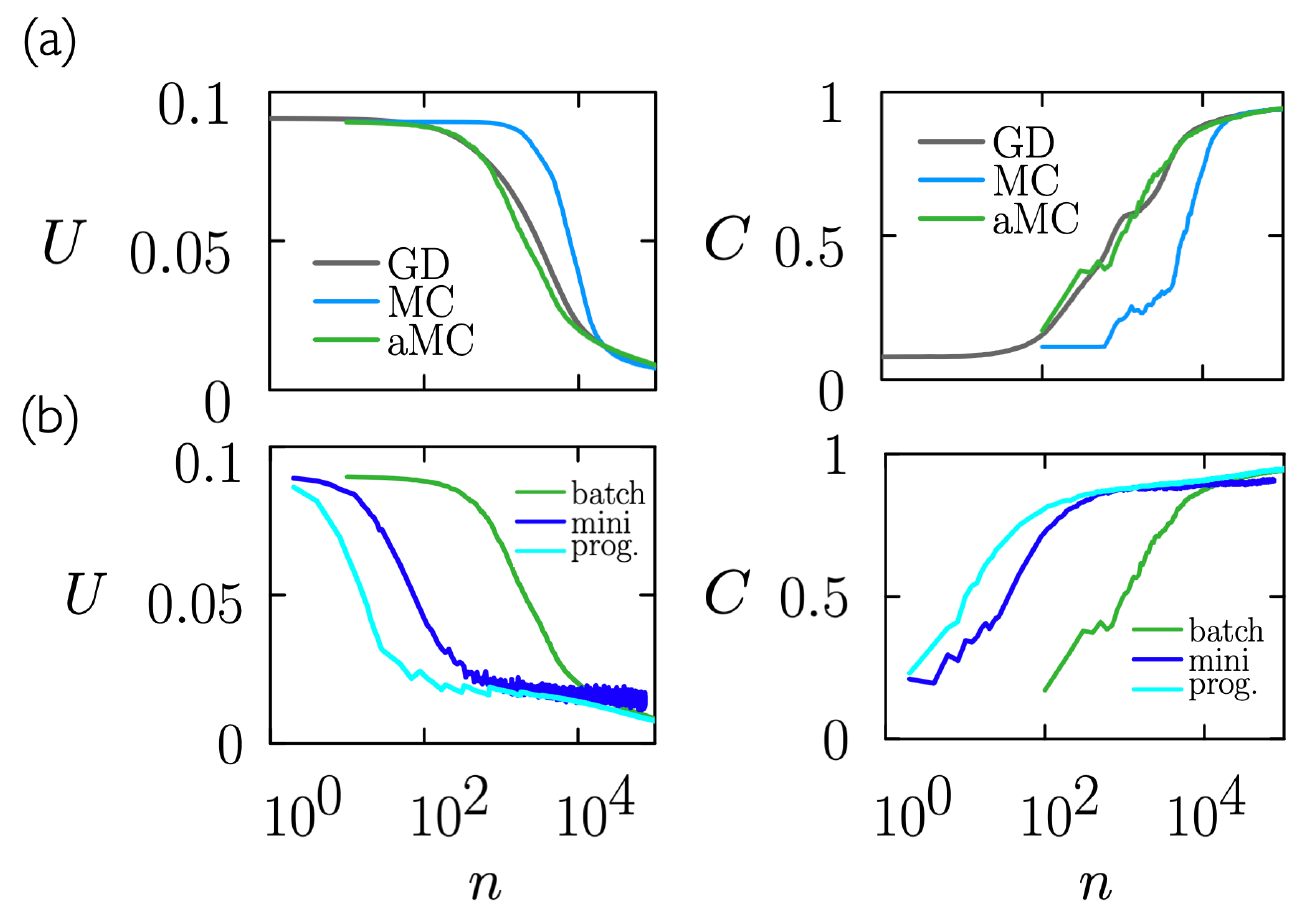} 
   \caption{As for gradient-based methods, different forms of batching can speed neural-net training by MC. (a) As \f{fig_zero}(b), with the addition of a simulation done using aMC (green) with hyperparameters $(\sigma_0,\epsilon, n_{\rm s},{\rm s.n.})=(10^{-2},0,20,{\rm on})$. (b) The aMC result from panel (a), which uses batch learning, compared with aMC using conventional minibatching (blue) and progressive batching (cyan).}
   \label{fig_comparison}
\end{figure}

\begin{figure}[] 
   \centering
   \includegraphics[width=\linewidth]{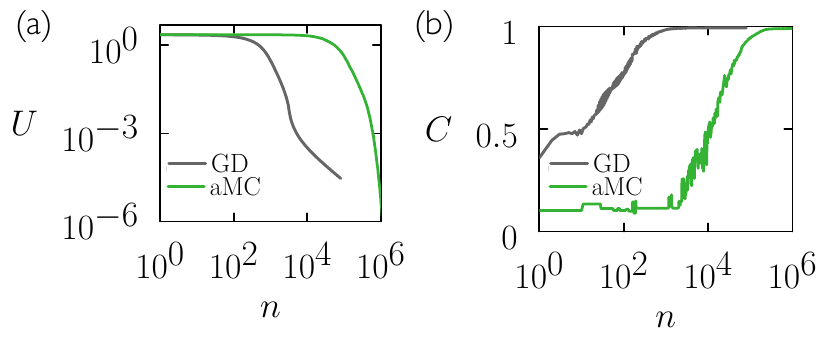} 
   \caption{Training-set loss $U$ (a) and test-set classification accuracy $C$ (b) for MNIST classification done using ResNET-18, a deep neural network of $\approx 1.11 \times 10^7$ parameters. GD learning rate was $10^{-3}$; aMC hyperparameters were  $\sigma_0 = 10^{-3}, n_s =100$, and $\epsilon = 10^{-3}$, with signal norm off.}
   \label{fig_resnet}
\end{figure}

In this section we first revisit \f{fig_zero} using aMC, and we present data indicating that numerical best practices for training nets using MC may differ from those developed for gradient-based algorithms.

In \f{fig_comparison}(a) we reproduce \f{fig_zero}(b) with the addition of an aMC simulation (green) in which we use aMC's adaptive step-size attenuation (with $n_{\rm s}=20$) and signal norm. These features allow us to choose an initial step-size parameter $\sigma_0=10^{-2}$ larger than the optimum value for Metropolis MC (see \f{fig_zero}(a)). As a result, training proceeds faster than for MC, at a rate comparable to the GD result shown. Training-set loss and test-set accuracy at long times are similar for all three algorithms. 

In \f{fig_comparison}(b) we compare the aMC result of panel (a), which uses batch learning (green), with two additional aMC results. The first (cyan) uses conventional minibatching with a minibatch size of 2000. Minibatch learning proceeds faster than batch learning, as happens with gradient-based methods, but achieves slightly larger values of training-set loss and test-set error than does batch learning. We speculate that this happens because of competing sources of stochasticity, that of the minibatch and that intrinsic to the MC algorithm. The second (cyan) uses progressive batching: training begins with a minibatch of size 500, which doubles every time the classification error rate on the minibatch falls below $10\%$ (aMC moves are conditioned against minibatch loss, in the usual way, not minibatch classification error, but the latter is the trigger for the doubling of the minibatch size). After doubling the minibatch size the aMC algorithm is reset ($\sigma \to \sigma_0$ and $\mu_i \to 0$). Progressive-batch learning proceeds faster than batch learning, and reaches similar final values of training-set loss and test-set accuracy. This comparison suggests that MC may respond differently than GD to procedures such as minibatch training; best practices for MC training of neural networks await development.

We note that the noise intrinsic to the Monte Carlo method provides a means of exploration even when the batch identity is kept fixed (noise also provides a way to effect change in the presence of vanishing gradients; see \s{deep} and \s{rnn}). For the problem discussed in this section there is a variation of about 1\% in values of test-set accuracy at $10^5$ epochs for 30 independent MC trajectories propagated with the same set of hyperparameters. Such fluctuations could provide the basis for an additional form of importance sampling that identified and propagated the best-performing networks in a population.

Finally, we show in \f{fig_resnet} the analog of \f{fig_zero}(b) for ResNET-18. ResNET-18 is a large, deep neural network with $\approx 1.11 \times 10^7$ parameters\c{he2016deep} (we changed the number of input channels for the first layer in order to apply it to MNIST). We applied layer normalization\c{ba2016layer}, which homogenizes neuron inputs and removes (or reduces) the need for signal norm when using aMC. The gradient descent learning rate was set to $10^{-3}$ and we applied gradient clipping for $|\nabla U|>1$; the aMC hyperparameters were $\sigma_0 = 10^{-3}, n_s =100$, and $\epsilon = 10^{-3}$. Two points are apparent from the plot: gradient descent trains faster than aMC, but aMC has similar ability to train in the long-time limit: both GD and aMC train to small loss and about $99.2\%$ accuracy on the MNIST test set. Thus for large modern architectures such as ResNET-18 we find training with gradients faster than training by Monte Carlo, but the latter has similar capacity for learning, suggesting that it is a promising tool for training neural networks when gradients are unreliable: see \s{rnn} and \s{deep}.

\section{Summary of aMC}
\label{amc}

\subsection{An adaptive version of the Metropolis algorithm for training neural networks}
\label{amc1}

In this section we summarize aMC, the adaptive Monte Carlo algorithm used in this paper. It is based on the Metropolis MC algorithm, modified to allow the move-proposal distribution to adapt in response to accepted and rejected moves. The Metropolis acceptance criterion is $\min(1,{\rm e}^{-\Delta U/T})$, where $\Delta U$ is the change of loss and $T$ is temperature. For nonzero temperature the algorithm allows moves uphill in loss. We focus here on the limit of zero temperature, which allows no uphill moves in loss. This choice is motivated by the success of gradient-descent algorithms and the intuition in deep learning (suggested by the structure of high-dimensional Gaussian random surfaces) that at large loss most stationary points on the loss surface are saddle points that can be escaped by moving downhill\c{dauphin2014identifying,bahri2020statistical}.

aMC is specified by four hyperparameters: $\sigma_0$, the initial move scale; $\epsilon$, the rate at which the mean of the move-proposal distribution is modified; $n_{\rm s}$, the number of consecutive rejected moves allowed before rescaling the parameters of the move-proposal distribution; and by the choice of signal norm being on or off. 

We introduce a counter $n_{\rm cr}=0$ to record the number of consecutive rejected moves. We initialize the parameters (weights and biases) $\x=\{x_1,\dots,x_i,\dots,x_N\}$ of the neural network (e.g. using Gaussian random numbers $x_i \sim {\cal N}(0 ,\sigma_0^2)$), and set the centers $\mu_i$ of each parameter's move-proposal distribution to zero. aMC proceeds as follows. 

\begin{enumerate}
\item[1.] {\em Current state.} Record the current neural-network parameter set $\x$. Select the data (defining the batch, episode, etc.) and record the current value of the loss $U(\x)$ on the data (for batch learning the value $U(\x)$ is known from the previous step of the algorithm). If signal norm is on, calculate the values $\lambda_i$ specified by \eqq{fan_in}, the required quantities having been calculated in the course of calculating $U(\x)$.

\item[2.] {\em Proposed move.} Propose a change 
\beq
\label{proposal}
x_i \to x_i'= x_i + \epsilon_i \quad {\rm with} \quad \epsilon_i \sim {\cal N}(\mu_i, \sigma_i^2)\eeq
of each neural-network parameter $i$, where $\sigma_i = \lambda_i \sigma$. Initially, $\mu_i=0$ and $\sigma = \sigma_0$, where $\sigma_0$ is the initial move scale. The parameters $\lambda_i$ are set either to unity (``signal norm off'') or by \eqq{fan_in}  (``signal norm on''). Evaluate the loss $U(\x')$ at the set of coordinates $\x'$ resulting from the proposal \eq{proposal}. If $U(\x') \leq U(\x)$~\footnote{For finite temperature $T$ the move is accepted if $\xi < {\rm e}^{(U(\x)-U(\x'))/T}$, where $\xi$ is a random number drawn uniformly on $(0,1]$.} then we accept the move and go to Step 3. Otherwise we reject the move and go to Step 4.
\item [3.] {\em Accept move.} Make the proposed coordinates $\x'$ the current coordinates $\x$. Set $n_{\rm cr} = 0$. For each neural-network parameter $i$, set 
\beq
\mu_i \to \mu_i + \epsilon(\epsilon_i-\mu_i)
\eeq
using the values $\epsilon_i$ calculated in \eq{proposal}.
Return to Step 1.
\item [4.] {\em Reject move.} Retain the set of coordinates $\x$ recorded in Step 1. Set $n_{\rm cr}\to n_{\rm cr}+1$. If $n_{\rm cr}=n_{\rm s}$ then set $n_{\rm cr} = 0$, $\sigma \to 0.95 \sigma$, and (for all $i$) $\mu_i = 0$. Return to Step 1.
\end{enumerate}

The computational cost of one move is the cost to draw $N$ Gaussian random numbers and to calculate the loss function twice (once for batch learning). The memory cost is the cost to hold two versions of the model in memory, and (if $\epsilon \neq 0$ and signal norm is on) the values $\mu_i$ and $\lambda_i$ for each neural-net parameter. Note that the algorithm requires calculation of the loss $U(\x)$ only, and not of gradients of the loss with respect to the net parameters. 

We refer to this algorithm as aMC, for adaptive Monte Carlo (the term ``adaptive Metropolis algorithm'' has been used in a different context\c{rosenthal2011optimal}). Standard zero-temperature Metropolis Monte Carlo is recovered in the limit $\epsilon=0, n_{\rm s} = \infty$, and $\lambda_i=1$.

\subsection{Signal norm: enacting heterogenous weight updates in order to keep roughly constant the change of neuron inputs}
\label{sn}

The proposal step \eq{proposal} contains the parameter step size $\sigma_i = \lambda_i \sigma$. For some applications, particularly involving deep or heterogeneous networks, it is useful to choose the $\lambda_i$ in order to keep the scale of updates for each neuron approximately equal, following ideas applied to gradient-based methods\c{lecun1996effiicient}. We call this concept {\em signal norm}; when signal norm is off, all $\lambda_i=1$. When it is on, we proceed as follows.

Consider the class of neural networks for which the input to neuron $j$ (its pre-activation) is
\beq
I_j^\alpha = \sum_{i \to j}^{N_j} x_i S_i^\alpha,
\eeq
where the sum runs over all weights $x_i$ feeding into neuron $j$; $N_j$ is the fan-in of $j$ (the number of connections entering $j$); and $S_i^\alpha$ is the output of neuron $i$ (the neuron that the weight $x_i$ connects to neuron $j$) given one particular evaluation $\alpha$ of the neural network. Under the proposal \eq{proposal} the change of input to neuron $j$ is approximately~\footnote{This approximation assumes that the output neurons do not change under the move. This is not true, but the intent here is to set the basic move scale, and absolute precision is not necessary.}
\beq
\Delta_j^\alpha = \sum_{i \to j}^{N_j} \epsilon_i S_i^\alpha.
\eeq
We therefore have
\beq
\av{\Delta_j^\alpha} = \sum_{i \to j}^{N_j} \mu_i S_i^\alpha
\eeq
and 
\beq
\av{(\Delta_j^\alpha)^2} = \sum_{i \to j}^{N_j} \sum_{k \to j}^{N_j}\left[\mu_i \mu_k (1-\delta_{ik}) + (\sigma_i^2+\mu_i^2) \delta_{ik} \right]S_i^\alpha S_k^\alpha,
\eeq
where $\av{\cdot}$ is the expectation over the move-proposal distribution \eq{proposal}, and $\delta_{ik}$ is the Kronecker delta. The expected approximate variance of the change of input to neuron $j$ under the move \eq{proposal} is therefore
\beq
\av{(\Delta_j^\alpha)^2}-\av{\Delta_j^\alpha}^2=\sigma^2 \sum_{i \to j}^{N_j} \lambda_i^2 (S_i^\alpha)^2.
\eeq
This quantity, averaged over all $N_{\rm data}$ neural-net calls required to calculate the loss, is
\beq
\label{av_data}
[\av{(\Delta_j^\alpha)^2}-\av{\Delta_j^\alpha}^2]_{\rm data} = \sigma^2 N_{\rm data}^{-1} \sum_{\alpha=1}^{N_{\rm data}} \sum_{i \to j}^{N_j} \lambda_i^2  (S_i^\alpha)^2.
\eeq
We can choose the values of the $\lambda_i$ in order to ensure that the right-hand side of \eq{av_data} is always $\sigma^2$. A simple way to do so is to set equal the $\lambda_i$ for all weights $x_i$ feeding neuron $j$, in which case
\beq
\label{lam}
\lambda_i =  \left(N_{\rm data}^{-1} \sum_{\alpha=1}^{N_{\rm data}} \sum_{i' \to j}^{N_j} (S_{i'}^\alpha)^2\right)^{-1/2}.
\eeq
If all neuron outputs $S_{i'}^\alpha$ appearing in \eq{lam} vanish identically then the expression must be regularized; one option is to set $\lambda_i=0$ for weights feeding a neuron whose input neurons are zero for a given pass through the data.  Recall that the sum $\alpha$ runs over the input data; the sum $i' \to j$ runs over all neurons $i'$ whose connections feed $j$; $N_j$ is the fan-in of $j$ (the number of connections entering $j$); and $S_{i'}^\alpha$ is the output of neuron $i'$ given a particular evaluation $\alpha$ of the neural network. The values \eq{lam} can be calculated from the pass through the data immediately before the proposed move. 

Under \eq{lam}, weights on connections that feed into a neuron receiving many other connections will experience a smaller basic move scale than weights on connections that feed into a neuron receiving few connections. Similarly, weights on connections fed by active neurons will experience a smaller basic move scale than weights on connections fed by relatively inactive neurons.

Finally, if the parameter $x_i$ is a bias we choose $\lambda_i=1$.

To summarize, we consider two settings for the parameters $\lambda_i$ that set the step-size parameters $\sigma_i = \lambda_i \sigma$ in the proposal \eq{proposal}. The first setting, ``signal norm off'' (used in \f{fig_zero}, \f{fig_acc}(a,b), \f{fig_gen}, and \f{fig_rnn}), has $\lambda_i =1$ for all parameters $x_i$. 

The second setting, called ``signal norm on'' (used in \f{fig_acc}(c,d), \f{fig2}, and \f{fig_comparison}), has 
\beq
\label{fan_in}
\lambda_i=\begin{cases}
			1 & \text{if $x_i$ is a bias};\\
            {\rm \eqq{lam}} & \text{if $x_i$ is a weight into neuron $j$}.
		 \end{cases}
\eeq
 
 \section{Conclusions}
 
 We have examined the Metropolis Monte Carlo algorithm as a tool for training neural networks, and have introduced aMC, an adaptive variant of it. Monte Carlo methods are closely related to evolutionary algorithms, which are used to train neural networks\c{GA,GA2,montana1989training,salimans2017evolution}, but the latter are usually applied to populations of neural networks; the MC algorithms we have considered here are applied to populations of size 1, just as gradient descent is. For sufficiently small moves the Metropolis algorithm is effectively gradient descent in the presence of white noise\c{whitelam2021correspondence}. Thus on theoretical grounds the Metropolis algorithm should possess the ability to train a neural network to values of a loss function similar to those achieved by GD; this is indeed what we (and others\c{sexton1999beyond,rere2015simulated,tripathi2020rso}) have observed empirically, both for simple neural nets and for large, modern architectures. This correspondence does not guarantee similar training times, however, and we have found gradient-based methods to be faster in general, particularly for large and heterogenous neural nets. 
 
 aMC is an adaptive version of the Metropolis algorithm. The efficiency of aMC diminishes less quickly with decreasing loss and increasing net size than does the efficiency of the Metropolis algorithm, and aMC can train faster than Metropolis, much as adaptive gradient-based methods can train faster than pure gradient descent.
 
 The Metropolis algorithm and aMC offer a complement to gradient-based methods in that they can sense the gradient when it exists but can work without it. In particular, aMC can train nets in which the gradient is too small (or too large) to allow gradient-based methods to train on the timescales simulated. We have shown here that aMC can train deep neural networks and recurrent neural networks that gradient descent cannot train. In both cases there exist modifications to those networks that can be trained by gradient-based methods, but aMC does not require those modifications. The design principles of neural nets optimal for Monte Carlo algorithms are largely unexplored but are likely distinct from those optimal for gradient-based methods, and having both sets of algorithms offers more choices for net design than having only one.
 
Finally, we note that while Metropolis and aMC have a fundamental connection to gradient-based methods in the limit of small step size, Monte Carlo algorithms more generally can enact large-scale nonlocal or collective changes that cannot be made by integrating gradient-based equations of motion\c{swendsen1987nonuniversal,wolff1989collective,frenkel2001understanding,chen2001improving,liu2004rejection,whitelam2007avoiding}.  The analogy suggests that improved Monte Carlo algorithms for training neural networks await development. 
  
\section{Code availability} Calculations using gradient descent and Adam were done in PyTorch\c{paszke2019pytorch}. Monte Carlo calculations were done in C and in PyTorch. A PyTorch implementation of the aMC optimizer (with signal norm off) and an example RNN optimization are available here\c{github}.

\section{Acknowledgments} This work was performed as part of a user project at the Molecular Foundry, Lawrence Berkeley National Laboratory, supported by the Office of Science, Office of Basic Energy Sciences, of the U.S. Department of Energy under Contract No. DE-AC02--05CH11231. This work used resources of the National Energy Research Scientific Computing Center (NERSC), a U.S. Department of Energy Office of Science User Facility operated under Contract No. DE-AC02-05CH11231. I.T. acknowledges funding from the National Science and Engineering Council of Canada. C.C. acknowledges a mobility grant from Research Foundation -- Flanders (FWO).


%

\end{document}